\documentclass[lettersize,journal]{IEEEtran}
\usepackage{amsmath,amsfonts}
\usepackage{algorithmic}
\usepackage{algorithm}
\usepackage{array}
\usepackage[caption=false,font=normalsize,labelfont=sf,textfont=sf]{subfig}
\usepackage{textcomp}
\usepackage{stfloats}
\usepackage{url}
\usepackage{verbatim}
\usepackage{graphicx}
\usepackage{cite}
\usepackage{color}
\usepackage{cite}
\usepackage{amssymb,amsfonts}
\usepackage{hyperref}
\hypersetup{hidelinks=true}
\usepackage{multirow}
\usepackage{multicol}
\usepackage{float}
\usepackage{doi}

\DeclareMathOperator{\E}{\mathbb{E}}
\def\BibTeX{{\rm B\kern-.05em{\sc i\kern-.025em b}\kern-.08em
    T\kern-.1667em\lower.7ex\hbox{E}\kern-.125emX}}

\begin{document}

\title{Phase-shifted Remote Photoplethysmography for Estimating Heart Rate and Blood Pressure from Facial Video}


\author{Gyutae Hwang \href{https://orcid.org/0000-0001-6365-2231}{\includegraphics[scale=0.3]{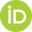}}, and Sang Jun Lee \href{https://orcid.org/0000-0002-9312-6299}{\includegraphics[scale=0.3]{ORCID-iD_icon_32x32.png}}
\thanks{This work was supported by the Institute of Information \& Communications Technology Planning \& Evaluation(IITP)-Innovative Human Resource Development for Local Intellectualization program grant funded by the Korea government(MSIT)(IITP-2024-RS-2024-00439292)(Corresponding author: Sang Jun Lee.)}
\thanks{Gyutae Hwang and Sang Jun Lee are with the Division of Electronics and Information Engineering, Jeonbuk National University, 567 Baekje-daero, Deokjin-gu, Jeonju, 54896, Jeonbuk-do, Republic of Korea (e-mail: gyutae741@jbnu.ac.kr; sj.lee@jbnu.ac.kr).}}



\maketitle

\begin{abstract}
Human health can be critically affected by cardiovascular diseases, such as hypertension, arrhythmias, and stroke.
Heart rate and blood pressure are important physiological information for monitoring of cardiovascular system and early diagnosis of cardiovascular diseases.
Previous methods for estimating heart rate are mainly based on electrocardiography and photoplethysmography, which require contacting sensors to skin surfaces.
Existing cuff-based methods for measuring blood pressure cause inconvenience and are difficult to be utilized in daily life.
To address these limitations, this paper proposes a two-stage deep learning framework for estimating heart rate and blood pressure from facial video.
The proposed algorithm consists of a dual remote photoplethysmography network (DRP-Net) and bounded blood pressure network (BBP-Net).
DRP-Net infers remote photoplethysmography (rPPG) signals at acral and facial sites, and these phase-shifted rPPG signals are utilized to estimate heart rate.
BBP-Net integrates temporal features and analyzes phase discrepancy between the acral and facial rPPG signals to estimate systolic blood pressure and diastolic blood pressure.
We augmented facial videos in temporal aspects by utilizing a frame interpolation model to increase bradycardia and tachycardia data.
Moreover, we reduced blood pressure error by incorporating a scaled sigmoid function in the BBP-Net.
Experiments were conducted on MMSE-HR and V4V datasets to demonstrate the effectiveness of the proposed method.
Our method achieved the state-of-the-art performance for estimating heart rate and blood pressure with significant margins compared to previous methods.
Our code is available at \href{https://github.com/GyutaeHwang/phase_shifted_rPPG}{https://github.com/GyutaeHwang/phase\_shifted\_rPPG}.
\end{abstract}


\section{Introduction}
\label{introduction}
\IEEEPARstart{T}{he} cardiovascular system consists of heart and blood vessels, and it circulates blood throughout the body, delivering oxygen and nutrients to tissues while removing waste substances.
Abnormalities in the cardiovascular system can lead to various cardiovascular diseases such as cardiomyopathy, hypertensive heart disease, arrhythmias, with potentially severe implications for overall health.
Moreover, cardiovascular diseases associated with blood vessels, such as hypertension and stroke, contribute to increasing the global mortality rate~\cite{gaidai2023global}.
However, these diseases can be detected at an early stage through the monitoring of heart rate (HR) and blood pressure (BP) using healthcare devices such as smartwatches.
For example, arrhythmias caused by disorders of the sinoatrial node can be prevented through the measurement of bradycardia, tachyarrhythmia, and irregular heart rates.
Additionally, systolic blood pressure (SBP) above 120 mmHg and diastolic blood pressure (DBP) above 90 mmHg may indicate the potential presence of hypertension.
Recently, interest in cardiovascular diseases has led to growing attention on research for healthcare services~\cite{aggarwal2023early,faust2022heart} and monitoring of physiological information~\cite{guo2022wrist,gupta2023support}.

Heart rate and blood pressure are main physiological information for the monitoring of cardiovascular diseases~\cite{perret2009heart,fuchs2020high}, and it can be measured by utilizing various physiological sensors and medical equipment.
Heart rate is commonly measured by utilizing electrocardiography (ECG) and photoplethysmography (PPG) sensors.
ECG employs electrodes attached to body to record electrical activities caused by contractions and relaxations of the heart.
PPG is a non-invasive method that uses a light source and photodetector attached at skin to measure volumetric variations of blood in microvessels.
The heart rate can be measured by computing peak-to-peak intervals of physiological signals in the time domain or by analyzing power spectrum in the frequency domain.
On the other hand, blood pressure can be measured based on oscillometric methods, which record the magnitude of oscillations using a blood pressure cuff.
A catheter-based method is an invasive approach to measure blood pressure, and it involves direct insertion of a sensor into an artery to measure real-time arterial blood pressure (ABP).
SBP and DBP can be computed from peak and valley values of the ABP signals.
Pulse transit time (PTT) is temporal delay of blood pulse waves which travel from the heart to an acral site such as a fingertip, and it is known that PTT is closely correlated with blood pressure~\cite{geddes1981pulse}.
PTT is measured differently depending on the distance between the heart and acral sites, resulting in a temporal delay in the PPG signals at each region.
Although there have been proposed blood pressure estimation methods by analyzing PTT from ECG or PPG signals~\cite{barvik2021noninvasive,ganti2020wearable}, these approaches have intrinsic limitations of requiring skin contacts.

Recently, camera sensors have been utilized to obtain physiological signals through a contactless method called remote photoplethysmography (rPPG).
From a facial video, rPPG technique extracts subtle variations in skin color induced by cardiac pulses.
The extracted temporal changes in pixel intensities are then transformed into continuous waveforms analogous to conventional PPG signals.
The precision of rPPG techniques can be critically affected by many factors such as light conditions, motion artifacts, and different skin tones.
Despite these challenges, rPPG has received attention due to non-contact and non-invasive attributes that facilitate remote monitoring of physiological information.

Deep learning methods have been utilized to extract rPPG signals and to estimate heart rate~\cite{chen2018deepphys,yu2022physformer,ouzar2023x}.
Additionally, various datasets have been released for the development and evaluation of deep learning algorithms~\cite{bobbia2019unsupervised, niu2019vipl, zhang2016multimodal, revanur2021first}.
During collecting the datasets, PPG signals were usually measured at acral sites such as fingertips, and therefore, there exists temporal discrepancy between PPG signals and the corresponding facial videos.
However, most previous methods neglected temporal discrepancy between the two modalities; time domain losses have been utilized to measure the difference between PPG signals at acral sites and rPPG signals at facial regions. 
For example, Lu et al.~\cite{lu2021dual} and Yu et al.~\cite{yu2022physformer} employed negative Pearson correlation loss to measure the similarity between PPG signals and estimated rPPG signals.
In~\cite{speth2024mspm}, the authors constructed a multi-site physiological monitoring (MSPM) dataset consisting of full-body PPG signals, verifying the differences in PTT across various body sites.
Moreover, Dong et al.~\cite{dong2024realistic} found significant trends and phase differences between facial and acral PPG by constructing a contact PPG acquisition system.
In this paper, we introduce the concept of phase-shifted rPPG signals, consisting of facial rPPG and acral rPPG signals, to analyze the temporal discrepancy between facial videos and PPG signals.
The acral rPPG signal is guided by a time domain loss function, while the facial rPPG signal is guided by frequency domain features which have temporally global characteristics.

On the other hand, blood pressure has been estimated by utilizing PPG signals, multimodal physiological signals, or facial videos.
While PPG-based methods have been widely employed in wearable devices, this approach has limitations of requiring physical contact to obtain physiological signals.
Although multimodal sensors which measure PPG and ECG signals can accurately estimate blood pressure by analyzing PTT, it also has the disadvantage of requiring contact with ECG and PPG sensors.
Recently, camera-based methods have been proposed to estimate blood pressure in a non-contact manner by utilizing spatiotemporal features in facial videos.
However, there is a performance gap between the camera-based and PPG-based methods.
\cite{huang2024camera, hou2025exploiting} leverages PTT from different facial regions to extract features related to blood pressure, yet there is a lack of experimental results on publicly available datasets.
The objective of this study is to improve the performance of the camera-based approach by utilizing phase-shifted information in acral and facial rPPG signals.

This paper proposes a two-stage deep learning pipeline consisting of a dual remote photoplethysmography network (DRP-Net) and bounded blood pressure network (BBP-Net).
In the first stage, DRP-Net infers acral and facial rPPG signals, and these signals are utilized to estimate heart rate.
In the second stage, BBP-Net analyzes phase discrepancy between the acral and facial rPPG signals and integrates temporal features based on a multi-scale fusion (MSF) module to estimate SBP and DBP values.
A scaled sigmoid function is employed in the BBP-Net to improve the precision of blood pressure estimation by constraining the estimated values into a predefined range.
Experiments were conducted on the MMSE-HR database~\cite{zhang2016multimodal} and V4V database~\cite{revanur2021first} to demonstrate the effectiveness of the proposed method for estimating heart rate and blood pressure from facial videos.
The main contributions of this paper can be summarized as follows.
\begin{itemize}
    \item {We propose a novel two-stage deep learning framework consisting of DRP-Net and BBP-Net for estimating heart rate and blood pressure.}
    \item {We introduce the concept of phase-shifted rPPG signals for extracting temporal discrepancy between pulse waves in facial videos and PPG signals measured at acral sites.}
    \item {We employ a frame interpolation algorithm for temporal augmentation of video clips to generate bradycardia and tachycardia data.}
    \item{We propose a novel loss function for the training of the DRP-Net and introduce a scaled sigmoid layer on the BBP-Net to improve the accuracy of estimating heart rate and blood pressure.}
    \item{Our proposed method achieved the state-of-the-art performance for estimating heart rate and blood pressure on the MMSE-HR and V4V datasets.}
\end{itemize}
The rest of this paper is organized as follows.
Section~\ref{sec:related} presents related work.
Section~\ref{sec:proposed} explains the proposed method for estimating phase-shifted rPPG signals, heart rate, and blood pressure.
Section~\ref{sec:exp} and Section~\ref{sec:conclusion} present experimental results and conclusion.

\section{Related work}
\label{sec:related}

\subsection{Estimation of rPPG signals and heart rate}
Conventional methods for extracting rPPG signals analyze subtle variations in color intensities of facial regions using image processing and mathematical modeling.
Poh et al.~\cite{poh2010advancements} computed pulse waves by averaging RGB channels, and independent component analysis was conducted to extract rPPG signals.
On the other hand, Lewandowska et al.~\cite{lewandowska2011measuring} proposed a channel selection process and estimated rPPG signals by conducting principal component analysis on color intensities of forehead regions.
De Haan \& Jeanne~\cite{de2013robust} proposed a chrominance-based method, called CHROM, and it improves the performance of extracting rPPG signals by minimizing the effect of motion artifacts.
Wang et al.~\cite{wang2016algorithmic} proposed the POS algorithm, and it analyzes a projection plane orthogonal to the skin tone in the normalized RGB space.
While these conventional methods are computationally efficient to extract rPPG signals, their performance is not sufficient to be utilized in real-world applications.

Recently, deep learning methods have been proposed to extract rPPG signals from facial videos.
Convolutional neural networks (CNNs) and transformer architectures have been utilized to analyze spatiotemporal features from image sequences.
Face detection and spatial attention modules are optionally utilized to improve the robustness to motion artifacts and external brightness conditions.
Chen \& McDuff~\cite{chen2018deepphys} introduced the convolutional attention network (CAN) that employs a Siamese-structured convolutional neural network. 
This model takes an image frame and the difference map to its adjacent frame, and the spatial attention module analyzes color variations in skin regions.
Nowara et al.~\cite{nowara2021benefit} introduced an inverse attention module to estimate corrupted signals affected by motion and illumination changes.
They further employed Long Short-Term Memory (LSTM) to enhance temporal robustness in estimating physiological signals.
Yu et al.~\cite{yu2019remote} proposed PhysNet3D, which consists of 3D CNN layers for extracting spatiotemporal features and deconvolution layers for recovering temporal details.
The PhysNet3D was trained by utilizing the Pearson correlation coefficient loss between PPG and estimated rPPG signals.
However, learning rPPG requires a rich temporal representation, which can be a weakness for CNN-based models with limited long-term dependency.
To address this issue, Yu et al.~\cite{yu2022physformer} proposed a video transformer consisting of temporal difference convolution (TDC) layers.
The TDC layers extract local spatial-temporal features to generate query and key projections, and multi-head self-attention mechanism is utilized to integrate global information.
Yu et al.~\cite{yu2023physformer++} further proposed a SlowFast Network to improve temporal representations of rPPG signals, and they demonstrated promising accuracy in estimating heart rate on cross-domain datasets.
Although transformer-based models enhance temporal representation with global features, they require high computational complexity.

\subsection{Deep learning methods for estimating blood pressure}

Deep learning models have been employed to analyze physiological signals to estimate blood pressure.
Miao et al.~\cite{miao2020continuous} proposed a deep learning model based on ResNet and LSTM to estimate continuous blood pressure from single channel ECG signals. 
Panwar et al.~\cite{panwar2020pp} introduced PP-Net, which consists of 1D convolution blocks and LSTM layers, to estimate SBP, DBP, and heart rate from PPG signals.
Huang et al.~\cite{huang2022mlp} analyzed PTT between PPG and ECG signals and employed MLP-Mixer to estimate blood pressure.
Moreover, Ma et al.~\cite{ma2024stp} proposed a data preprocessing method for transforming physiological features in PPG signals obtained from different sources to estimate blood pressure in self-supervised manner.

Recently, vision-based methods for estimating blood pressure have received much attention.
Most conventional approaches integrated rPPG methods to extract physiological signals and deep learning models to estimate blood pressure from PPG signals.
Wu et al.~\cite{wu2022facial} proposed FS-Net to estimate SBP and DBP values from three-channel rPPG signals and seven physiological indicators including heart rate and body mass index.
Bousefsaf et al.~\cite{bousefsaf2022estimation} employed continuous wavelet transform and a pre-trained U-Net model to estimate continuous BP signals from estimated rPPG signals.
The rPPG signals were obtained by spatially averaging green channel of skin regions in facial videos.
On the other hand, Chen et al.~\cite{chen2023remote} proposed an end-to-end network for estimating blood pressure from facial videos.
They extracted spatiotemporal features from four pre-defined facial regions and regressed SBP and DBP values by utilizing ResNet18 and bidirectional LSTM layers.
Previous studies typically extract visible pulse waves in a non-parametric manner and have difficulty fully utilizing the temporal features of facial videos.
To address this issue, we explore a deep learning-based phase-shifted rPPG estimation method as an intermediate step in BP estimation.
In Table~\ref{tab:related-works}, we summarize previous studies on deep learning-based heart rate and blood pressure estimation.

\begin{table}[]
    \caption{Summary of Deep Learning Models for HR and BP Estimation.}
    \label{tab:related-works}
    \resizebox{1\linewidth}{!}{%
    \setlength{\tabcolsep}{3pt}
    \renewcommand{\arraystretch}{1.5}
    \begin{tabular}{llll}
    \hline
    Methods                                        & Tasks  & Model architecture             & Input signal                               \\ \hline
    DeepPhys~\cite{chen2018deepphys}               & HR     & 2D CNN                         & Facial video                               \\
    Benefit of distraction~\cite{nowara2021benefit}& HR     & 2D CNN and bi-LSTM             & Facial video                               \\
    PhysNet~\cite{yu2019remote}                    & HR     & 3D CNN                         & Facial video                               \\
    PhysFormer~\cite{yu2022physformer}             & HR     & Video transformer              & Facial video                               \\
    PhysFormer++~\cite{yu2023physformer++}         & HR     & Video transformer              & Facial video                               \\
    Miao et al.~\cite{miao2020continuous}          & BP     & ResNet and LSTM                & ECG signal                                 \\
    PP-Net~\cite{panwar2020pp}                     & BP, HR & 1D CNN and LSTM                & PPG signal                                 \\
    MLP-BP~\cite{huang2022mlp}                     & BP     & MLP-Mixer                      & PPG and ECG signals                        \\
    SPT~\cite{ma2024stp}                           & BP     & Transformer                    & PPG signal                                 \\
    FS-Net~\cite{wu2022facial}                     & BP     & 2D CNN and FC layers           & \begin{tabular}[c]{@{}l@{}}rPPG signal and \\ 7 physiological indicators\end{tabular}\\
    Bousefsaf et al.~\cite{bousefsaf2022estimation}& BP     & 2D CNN                         & iPPG signal                                \\
    BPE-Net~\cite{chen2023remote}                  & BP     & 2D CNN and bi-LSTM             & Facial video                               \\ \hline
    \end{tabular}
    }
\end{table}

\section{Methodology}
\label{sec:proposed}

\subsection{Overall training pipeline}
\label{sec:overall}

\begin{figure*}[t]
    \centering
    \includegraphics[width=0.9\linewidth]{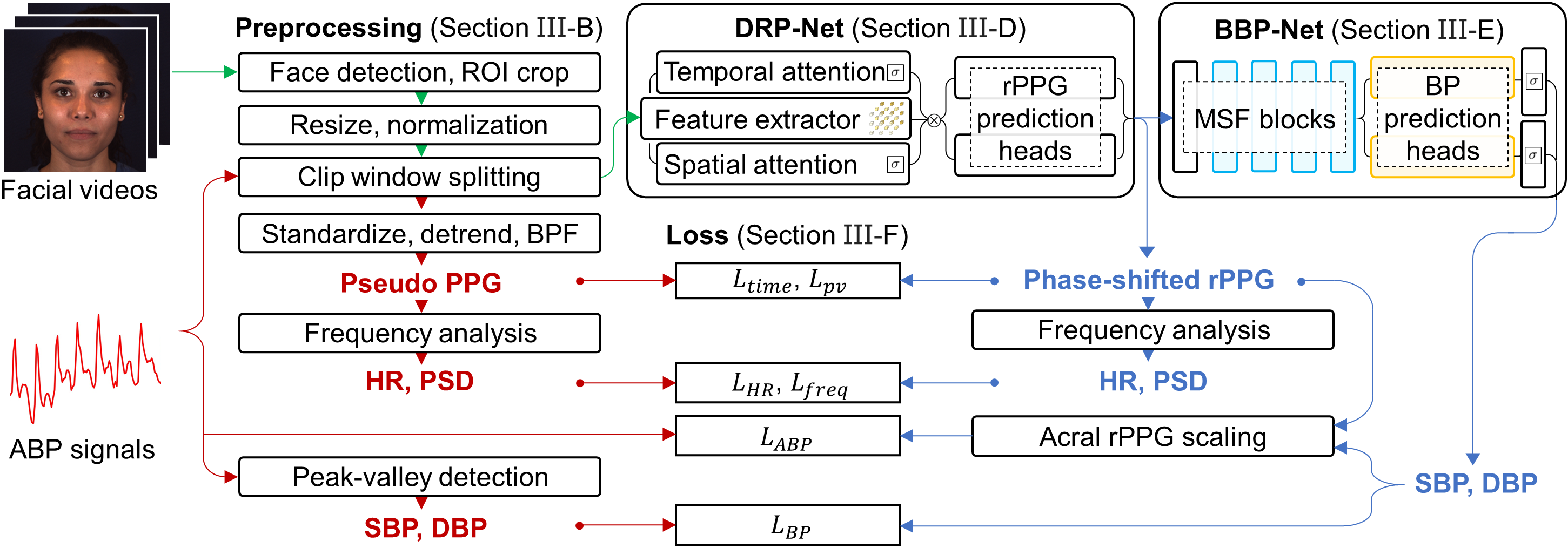}
    \caption{Overview of the training pipeline of the proposed method. Red and blue texts indicate ground truth and predicted physiological information, respectively. The green, red, and blue arrows represent the preprocessing of input videos, the generation of ground truth, and the post-processing of model outputs, respectively.}
    \label{fig:overview}
\end{figure*}

This paper proposes a two-stage deep learning framework consisting of DRP-Net and BBP-Net to estimate heart rate and blood pressure from facial videos.
The proposed deep learning model extracts acral and facial rPPG signals and analyzes their temporal discrepancy to estimate SBP and DBP.
Fig.~\ref{fig:overview} presents an overview of the training pipeline of the proposed method.
In the preprocessing step, facial regions are detected within a video clip to define a region of interest (ROI).
The DRP-Net learns spatiotemporal features from the ROI sequence and extracts phase-shifted rPPG signals, which consist of acral and facial rPPG signals.
The BBP-Net consists of MSF blocks and BP prediction heads, and it infers SBP and DBP values from the phase-shifted rPPG signals.
For ground truth generation, ABP signals are utilized to calculate pseudo PPG signals, heart rate, SBP, and DBP.

\begin{figure*}[t]
    \centering
    \includegraphics[width=\linewidth]{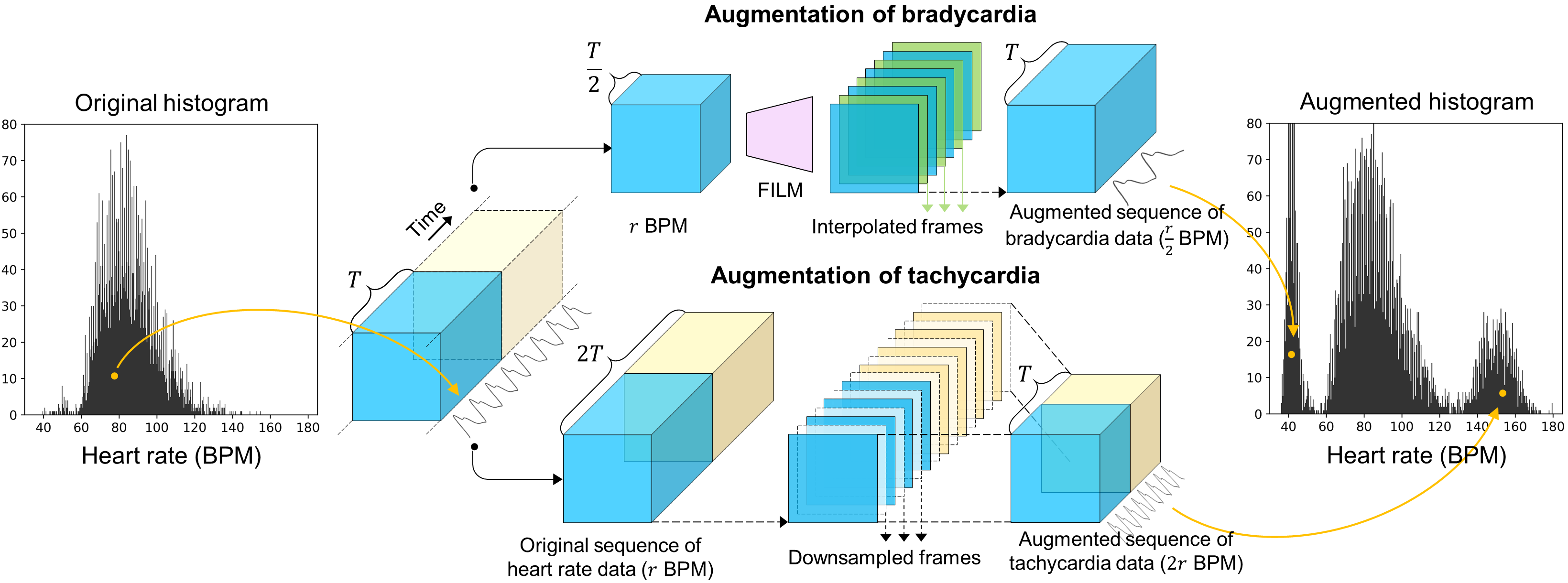}
    \caption{Data augmentation process of bradycardia and tachycardia samples.}
    \label{fig:augmentation}
\end{figure*}

\subsection{Preprocessing and ground truth generation}
\label{sec:pre}
The preprocessing step extracts normalized ROI regions from input images to remove redundant background information.
The pretrained MTCNN \cite{zhang2016joint} model is utilized to detect facial regions, and an ROI is decided to include facial regions within a short video clip.
Following DeepPhys~\cite{chen2018deepphys}, we use a fixed bounding box for each video with the scaling factor of 1.6 to address missed detections and handle subject movements.
The ROI regions are cropped and resized into the size of 128$\times$128, and their brightness is normalized into the range between 0 and 1 to reduce the effect of light conditions.
The cropped ROI sequence is sampled at 25 frames per second (FPS) and split into the window length of 150 frames which corresponds to 6 seconds following the previous method~\cite{chen2023remote}.

To generate the ground truth data, the ABP signals are synchronized with the facial videos, and they are sampled at 25 Hz.
The ABP signals are split into the window length of 6 seconds, and pseudo PPG signals are generated by conducting standardization, detrending, and bandpass filtering (BPF).
To generate pseudo PPG signals from ABP signals, the detrending algorithm proposed by Tarvainen et al.~\cite{tarvainen2002advanced} is employed for removing the effect of BP variations.
Following the previous work~\cite{nowara2021benefit}, BPF is conducted on physiological signals with a pre-defined range between 0.5 Hz and 3.0 Hz.
Fast Fourier Transform is conducted to compute the power spectral density (PSD) of pseudo PPG signals, and the heart rate is computed by analyzing the frequency corresponding to the maximum amplitude of the PSD.
Ground truth values for SBP and DBP are computed by averaging peak and valley values of ABP signals within each window.

\subsection{Data augmentation}
\label{sec:augmentation}
We utilized frame interpolation model to augment bradycardia and tachycardia data, and Fig.~\ref{fig:augmentation} presents the data augmentation process in temporal aspects.
Normal heart rates generally range from 60 BPM to 100 BPM~\cite{guyton2006text}, and bradycardia and tachycardia refer to heart rates lower and higher than the normal heart rate.
The left histogram in Fig.~\ref{fig:augmentation} shows the heart rate distribution of the V4V trainset. 
The limited heart rate data distribution can constrain the model's heart rate estimation range, triggering the need for data augmentation.
In the training process, an ROI sequence of the length $T$ corresponding to the heart rate $r$ beats per minute (BPM) is augmented to generate bradycardia and tachycardia data which correspond to the heart rates $\frac{r}{2}$ BPM and $2r$ BPM, respectively.
To augment the bradycardia data, $\frac{T}{2}$ frames of the original ROI sequence is interpolated into the length of $T$ frames by utilizing a pretrained FILM-Net~\cite{reda2022film}.
The duration of a cardiac cycle increases as the interpolation rate increases, resulting in decreased heart rate.
In contrast, $2T$ frames of the original ROI sequence is downsampled with the factor of $2$ to augment tachycardia data.
Previous data augmentation methods are based on frame sampling~\cite{yu2020autohr, lokendra2022and}, color jittering~\cite{yue2023facial}, and ROI masking~\cite{chen2023remote}.
Different to these previous approaches, data augmentation based on a frame interpolation model has the benefit of being able to directly control the heart rate of the augmented data.

\begin{figure*}[t]
    \centering
    \includegraphics[width=0.95\linewidth]{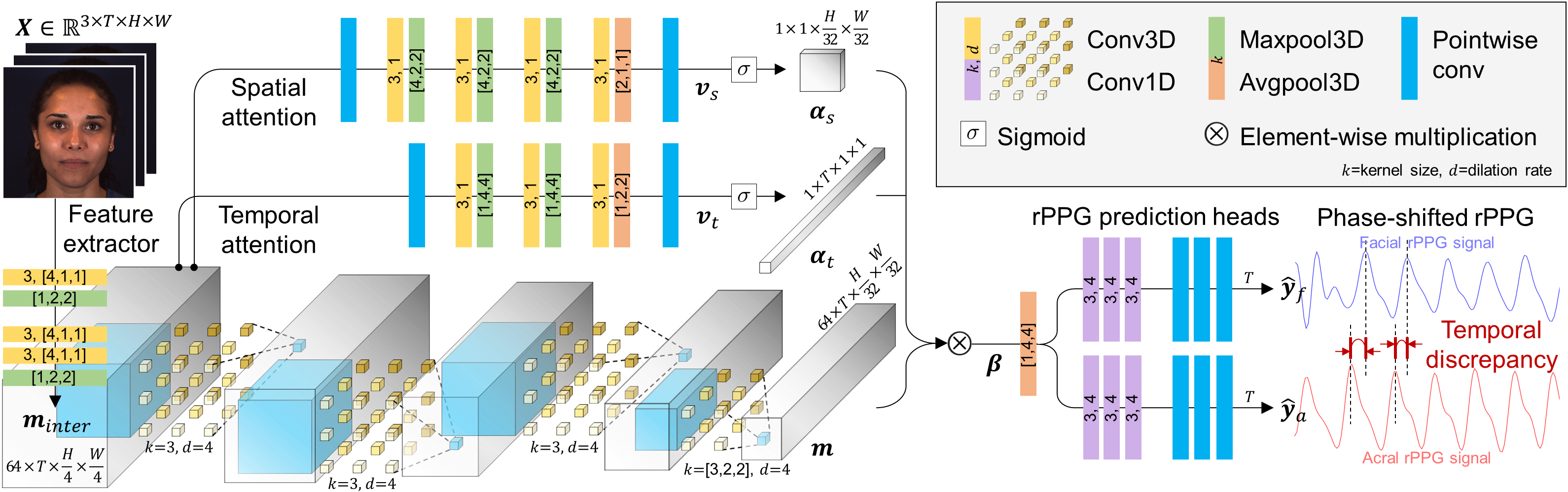}
    \caption{Architecture of DRP-Net.}
    \label{fig:DRP-Net}
\end{figure*}

\subsection{Dual remote photoplethysmography network (DRP-Net)}
\label{sec:DRP-Net}
In this paper, we propose DRP-Net to estimate phase-shifted rPPG signals from image sequences, and Fig.~\ref{fig:DRP-Net} presents the architecture of the DRP-Net.
We designed DRP-Net as a multi-task learning model to extract facial and acral rPPG signals that have similar trends but different phases.
DRP-Net takes a sequence of facial images $\textbf{X} \in \mathbb{R}^{3 \times T \times H \times W}$, where $H$ and $W$ are height and width, and $T$ is the number of frames in a window.
In experiments, $H$ and $W$ are set to 128, and $T$ is set to 150.
DRP-Net is a 3D CNN model, consisting of a feature extractor, spatial and temporal attention modules, and rPPG prediction heads.
The feature extractor consists of atrous convolution layers to analyze spatiotemporal features over a large receptive field.
In Fig.~\ref{fig:DRP-Net}, the kernel size and dilation ratio for each convolution layer are denoted by $k$ and $d$; in a 3D convolution layer, $k$ and $d$ are tuples with the form of [$T$, $H$, $W$], and it is represented as a scalar if three values are the same.
The feature extractor consists of seven atrous convolution layers and three max pooling layers, and it produces a spatiotemporal feature map $\textbf{m} \in \mathbb{R}^{64 \times T \times 4 \times 4}$.

The intermediate feature map $\textbf{m}_{inter} \in \mathbb{R}^{64 \times T \times 32 \times 32}$ of the feature extractor is utilized to compute spatial and temporal attention in parallel.
The objective of the spatial and temporal attention modules is to emphasize important regions of the face and the temporal peak locations within the facial video.
The spatial attention module consists of 3D convolution, max pooling, average pooling, and pointwise convolution layers.
The sigmoid function $\sigma$ is applied on the attention score map $\textbf{v}_s \in \mathbb{R}^{1 \times 1 \times 4 \times 4}$ to produce the spatial attention $\boldsymbol{\alpha}_s$.
Similarly, in the temporal attention module, we compute a temporal score vector $\textbf{v}_t \in \mathbb{R}^{1 \times T \times 1 \times 1}$, and the sigmoid function is applied to obtain the temporal attention vector $\boldsymbol{\alpha}_t$.
The spatial and temporal attention modules refine the spatiotemporal feature map $\textbf{m}$ as follows.
\begin{equation}
\label{eq3} 
\boldsymbol{\beta}=\textbf{m} \otimes \boldsymbol{\alpha}_t \otimes \boldsymbol{\alpha}_s,\end{equation}
where $\otimes$ denotes the operation of broadcasting and elementwise multiplication.

The rPPG prediction heads infer phase-shifted rPPG signals consisting of facial and acral rPPG signals, from the spatiotemporal feature map $\textbf{m}$.
The facial and acral rPPG signals are denoted as $\hat{\textbf{y}}_f \in \mathbb{R}^T$ and $\hat{\textbf{y}}_a \in \mathbb{R}^T$.
The rPPG prediction heads consist of a Siamese structure which contains 1D atrous convolution and pointwise convolution layers.
While two prediction heads have a same structure, they are trained by utilizing different loss functions to infer facial and acral rPPG signals.

\begin{figure*}[t]
    \centering
    \includegraphics[width=\linewidth]{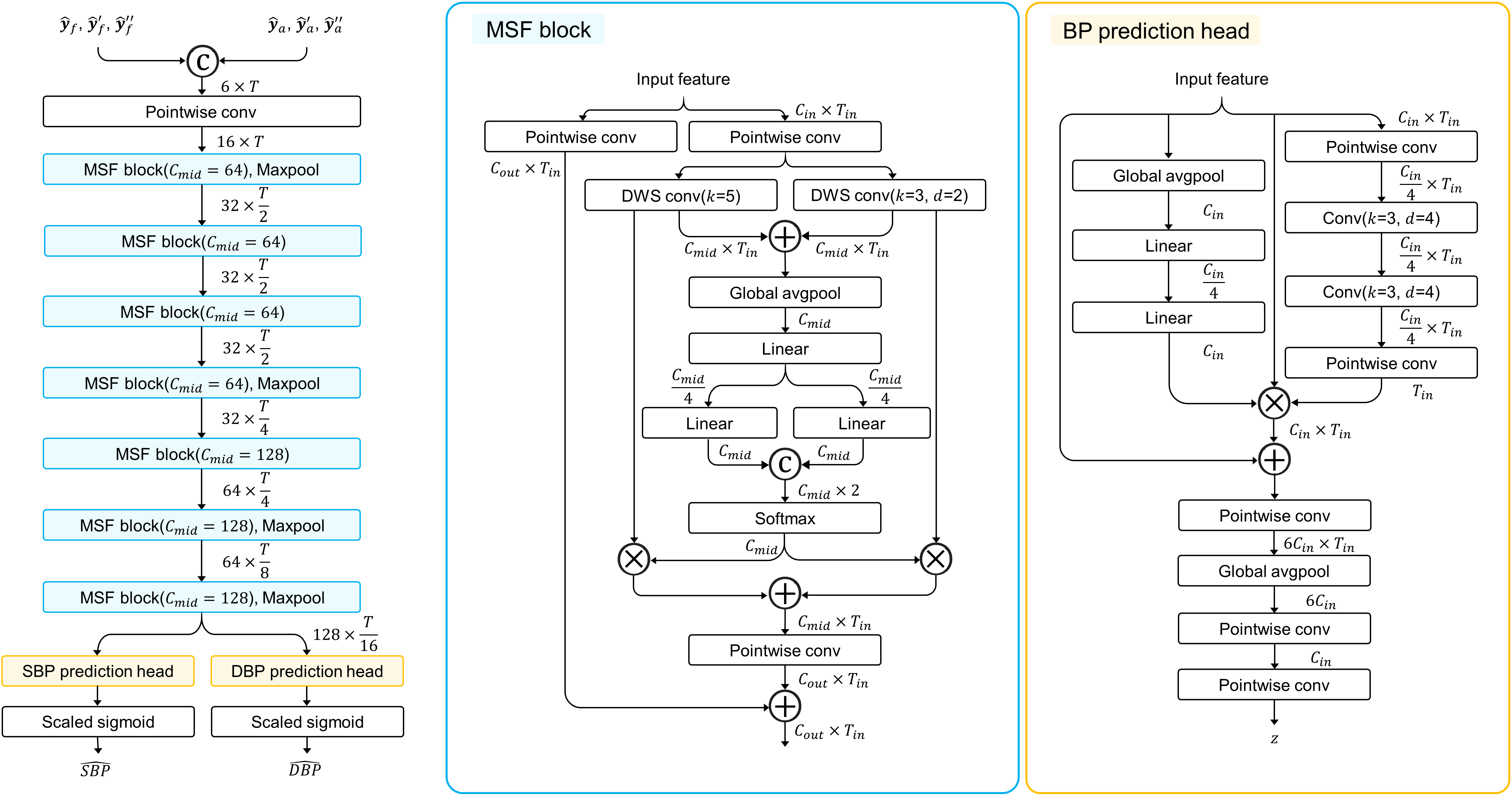}
    \caption{Architecture of BBP-Net.}
    \label{fig:BBP-Net}
\end{figure*}

\subsection{Bounded blood pressure network (BBP-Net)}
\label{sec:BBP-Net}
BBP-Net is designed to estimate SBP and DBP values from facial and acral rPPG signals, and its structure is presented in Fig.~\ref{fig:BBP-Net}.
BBP-Net takes a stack of physiological signals consisting of phase-shifted rPPG signals, velocity plethysmography (VPG) and acceleration plethysmography (APG) signals.
VPG and APG are the first and second derivatives of the facial and acral rPPG signals.
To extract local temporal features such as phase shifts and signal waveforms, BBP-Net is stacked with convolution-based modules, consisting of pointwise convolution layers, MSF blocks, BP prediction heads, and scaled sigmoid layers.

The MSF block integrates physiological information from various receptive fields.
In Fig.~\ref{fig:BBP-Net}, $C_{in}$, $C_{out}$, and $C_{mid}$ denote the channels of input, output, and middle layers in MSF blocks and BP prediction heads; $T_{in}$ denotes the temporal length of an input signal.
The MSF block performs depth-wise separable (DWS) convolution with the kernel sizes of 3 and 5 in parallel.
Since phase-shifted rPPG signals contain negative values, we employed the Hard Swish activation function [45] to handle negative values while preserving the smooth characteristics of the signal data.
After the global average pooling, the global feature vector with channel $C_{mid}$ passes linear layers and a softmax layer to compute weight vectors for the outputs of two DWS convolution layers.
MSF blocks contain a residual connection to reduce the problem of vanishing gradients.

The BP prediction head includes a bottleneck attention module~\cite{park2018bam} and residual connections to extract temporal features.
Moreover, we employed a scaled sigmoid function to constrain the estimated blood pressure into a predefined range as follows.
\begin{equation}
\label{eq4}
\widehat{BP}=BP_{min}+(BP_{max}-BP_{min})\sigma(z/\tau).
\end{equation}
In (\ref{eq4}), $\widehat{BP}$ is the estimated blood pressure, and ${BP}_{max}$ and ${BP}_{min}$ are heuristic parameters that represent upper and lower bounds of blood pressure.
The symbol $BP$ can be either $SBP$ and $DBP$, and the upper and lower bounds of $\widehat{SBP}$ are set to 155 mmHg and 85 mmHg.
On the other hand, the upper and lower bounds of $\widehat{DBP}$ are set to 95 mmHg and 45 mmHg.
In (\ref{eq4}), $\sigma$, $z$ and $\tau$ denote the sigmoid function, the output of the BP prediction head, and temperature, respectively.
We set the temperature parameter into 2 in experiments.
The structure of BBP-Net is based on the architecture proposed by Hu et al.~\cite{hu2022ppg}, which is designed to estimate SBP and DBP from PPG signals.
While the previous method takes a single PPG signal, our proposed model analyzes the phase discrepancy between the facial and acral rPPG signals and predicts more accurate BP by incorporating the scaled sigmoid function.

\subsection{Loss function}
\label{sec:loss}
The loss function for the training of DRP-Net consists of $L_{freq}$, $L_{HR}$, $L_{time}$, and $L_{pv}$.
The frequency domain loss $L_{freq}$ is computed from PSD of the physiological signals, and it is defined as
\begin{equation}
\label{eq6}
L_{freq}=\lVert P(\hat{\textbf{y}})-P(\textbf{y}) \lVert_{2},
\end{equation}
where $\hat{\textbf{y}}$ and $\textbf{y}$ denote an rPPG signal predicted by DRP-Net and the corresponding pseudo PPG signal obtained from ABP signals, respectively.
In (\ref{eq6}), $P(\cdot)$ indicates the operation for computing PSD of a physiological signal based on fast Fourier transform.
The PSD is analyzed within the frequency range between 0.5 Hz and 3 Hz, which corresponds to the heart rates between 30 BPM and 180 BPM.
A predicted heart rate is computed from the frequency corresponding to the maximum amplitude of the PSD.

$L_{HR}$ measures the absolute difference between the predicted heart rate and its ground truth heart rate, and it is defined as follows.
\begin{equation}
\label{eq5}
L_{HR}= |\widehat{HR}-HR|,
\end{equation}
where $\widehat{HR}$ and $HR$ are the predicted and ground truth heart rates, respectively.

In addition, we define a time domain loss $L_{time}$ to estimate the difference between the estimated acral rPPG signal $\hat{\textbf{y}}_a$ and its corresponding pseudo PPG signal $\textbf{y}$.
$L_{time}$ supervise the phase and morphological features of the acral rPPG signal by directly measuring the distance to its pseudo PPG signal based on (\ref{eq8}).
\begin{equation}\label{eq8} L_{time}=\lVert \hat{\textbf{y}}_a-\textbf{y}  \rVert_{2}.\end{equation}

We propose an additional time domain loss $L_{pv}$ to supervise the scale of the estimated rPPG signals.
Let $S_p(\textbf{y})$ and $S_v(\textbf{y})$ be the sets of time stamps corresponding to peak and valley values in a physiological signal $\textbf{y}$.
The set of time stamps corresponding to peak values is obtained as follows.
\begin{equation}
\begin{aligned}
    S_p(\textbf{y})= \{ t \mid (\textbf{y}(t)- \textbf{y}(t-1))( \textbf{y}(t+1)- \textbf{y}(t)) < 0,\\
    ~\textbf{y}(t) > \textbf{y}(t-1) \},
\end{aligned}
\end{equation}

where \textbf{y}(t) is the value of a physiological signal \textbf{y} at the time stamp $t$.
To remove dicrotic notches and noise components, $S_p(\textbf{y})$ is refined into $\tilde{S_p}(\textbf{y})$ as follows.
\begin{equation}
\tilde{S_p}(\textbf{y}) = \{t \mid t\in S_p(\textbf{y}), ~\textbf{y}(t) > \E_{t\in S_p(\textbf{y})} \left[ \textbf{y}(t) \right].
\end{equation}
Similarly, $S_v(\textbf{y})$ and $\tilde{S_v}(\textbf{y})$ are obtained based on (\ref{eq91}) and (\ref{eq92}).
\begin{equation}
\label{eq91}
\begin{aligned}
S_v(\textbf{y}) = \{ t \mid (\textbf{y}(t)- \textbf{y}(t-1))( \textbf{y}(t+1)- \textbf{y}(t)) < 0, \\
~\textbf{y}(t) < \textbf{y}(t-1) \}.
\end{aligned}
\end{equation}
\begin{equation}
\label{eq92}
\tilde{S_v}(\textbf{y}) = \{t \mid t\in S_v(\textbf{y}), ~\textbf{y}(t) < \E_{t\in S_v(\textbf{y})} \left[ \textbf{y}(t) \right] \}.
\end{equation}
The averaged peak and valley values are denoted as $p(\textbf{y})$ and $v(\textbf{y})$, and they are computed as follows.
\begin{equation}
p(\textbf{y}) = \E_{t\in \tilde{S_p}(\textbf{y})} \left[ \textbf{y}(t) \right].
\end{equation}
\begin{equation}
v(\textbf{y}) = \E_{t\in \tilde{S_v}(\textbf{y})} \left[ \textbf{y}(t) \right].
\end{equation}
The auxiliary loss function $L_{pv}$ is computed by measuring the L2 distance of averaged peak and valley values as follows.
\begin{equation}
L_{pv} = \sqrt{(p(\textbf{y}) - p(\hat{\textbf{y}}))^2 + (v(\textbf{y}) - v(\hat{\textbf{y}}))^2}.
\end{equation}

The total loss to optimize the facial rPPG signal is defined as \begin{equation}\label{eq9} 
L_{facial}=\lambda_1 L_{HR} + \lambda_2 L_{freq} + L_{pv},\end{equation}
and it guides the model to extract pulse waves which corresponds to the phase of the facial image sequence.
On the other hand, optimization of the acral rPPG signal utilizes the following loss function.
\begin{equation}\label{eq10} L_{acral}=\lambda_1 L_{HR} + \lambda_2 L_{freq} + L_{pv} + L_{time},\end{equation}
and it supervise rPPG signals to mimic the phase of pseudo PPG signals which are obtained from an acral site.
In experiments, the constants $\lambda_1$ and $\lambda_2$ are set to 0.0001 and 100, respectively.

For the training of BBP-Net, we define $L_{BP}$ and $L_{ABP}$.
$L_{BP}$ employs Huber loss~\cite{collins1976robust} to calculate the loss of predicted SBP and DBP, and it is defined as the following equation.
\begin{equation}\label{eq11} L_{BP}=
    \left\{\begin{matrix}
        \frac{1}{2}(\widehat{BP} - BP)^{2}, & if ~| \widehat{BP} - BP  | < \delta\\
        \delta   ( | \widehat{BP} - BP |  - \frac1 2 \delta ), & otherwise
    \end{matrix}\right.\end{equation} 
In (\ref{eq11}), $\widehat{BP}$ and $BP$ denote predicted and ground truth blood pressure, and the notation $BP$ can be either $SBP$ or $DBP$.
In experiments, the heuristic parameter $\delta$ is set to 1.
The Huber loss imposes quadratically increasing penalty within the pre-defined range $\delta$, and $L_{BP}$ increases linearly if the absolute difference is larger than $\delta$.

In addition, we define a time domain loss $L_{ABP}$ to reconstruct ABP signals based on the predicted physiological information.
The scaled version of the acral rPPG signal $\hat{\textbf{y}}_{a}$ is denoted as $\hat{\textbf{y}}_{s}$, and it can be computed as follows.
\begin{equation}\label{eq12} \hat{\textbf{y}}_{s}= \frac{\hat{\textbf{y}}_{a} - \hat{y}_{min}}{\hat{y}_{max}-\hat{y}_{min}}(\widehat{SBP} - \widehat{DBP}) + \widehat{DBP},\end{equation}
where $\hat{y}_{max}$ and $\hat{y}_{min}$ are the maximum and minimum values of $\hat{\textbf{y}}_{a}$.
$L_{ABP}$ is defined as the L2 distance between the scaled acral rPPG signal $\hat{\textbf{y}}_{s}$ and its corresponding ABP signal $\textbf{y}_{ABP}$ as follows.
\begin{equation}\label{eq13} L_{ABP}=\lVert \hat{\textbf{y}}_{s}-\textbf{y}_{ABP} \rVert_{2}.\end{equation}
$L_{ABP}$ supervise $\widehat{SBP}$ and $\widehat{DBP}$ to reduce the gap between the reconstructed and ground truth ABP signals.

\section{Experimental results}
\label{sec:exp}

Experiments were conducted on a hardware environment including Intel Core i9-10940X CPU, 64 GB DDR4 RAM, and NVIDIA Geforce RTX 3090 Ti.
Pytorch was utilized to implement the proposed algorithm, and our code is available at \href{https://github.com/GyutaeHwang/phase_shifted_rPPG}{https://github.com/GyutaeHwang/phase\_shifted\_rPPG}.
In experiments, the temporal window was set to 150 samples which corresponds to 6 seconds.
The learning rates of the Adam optimizer were set to 0.001 and 0.0001 for the MMSE-HR and V4V database, and the batch size was set to 8.
To evaluate the accuracy of estimated heart rate and blood pressure, we adopted the mean absolute error (MAE), root mean squared error (RMSE), and Pearson correlation coefficient \textit{r} as evaluation measures.

\subsection{Datasets}
We conducted experiments using the MMSE-HR (Multimodal Spontaneous Expression-Heart Rate)~\cite{zhang2016multimodal} and V4V (Vision for Vitals) database~\cite{revanur2021first}.
The datasets are sub-datasets derived from the MMSE database (BP4D+), which consists of synchronized facial image sequences and continuous ABP signals. 
Before collecting physiological data, each subject signed an informed consent form in accordance with the IRB approved protocol.
As explained in Section~\ref{sec:pre}, pseudo PPG signals, heart rate, SBP, and DBP were obtained from ABP signals.
In addition, the ABP measurement device used in these datasets is the Biopac NIBP100D, which can non-invasively measure continuous ABP signals by calibrating the finger PPG signals using cuff data.
RGB video was recorded at the resolution of 1040$\times$1392 with the frame rate of 25 FPS, and ABP signals were collected at the sampling rate of 1000 Hz.
The MMSE-HR database includes 102 video sequences from 17 male and 23 female subjects, and the average length of sequences is 30 seconds.
The V4V database includes 1,358 data sequences from 179 subjects, and they are split into 724 for training, 276 for validation, and 358 for test samples.
The average length of the sequences in the V4V database is 40 seconds.

During the acquisition of the MMSE database, subjects performed various tasks to arouse target emotions. The MMSE-HR dataset contains tasks designed to arouse emotions such as amusement, physical pain, anger, and disgust, while the V4V dataset additionally includes surprise, sadness, startle, skepticism, embarrassment, and fear. Specifically, tasks for surprise, startle, and embarrassment can induce large head motions, making the V4V dataset more challenging.

\subsection{Experimental results on MMSE-HR database}
The performance of the DRP-Net for estimating heart rate is compared with previous methods on the MMSE-HR dataset.
Following the previous work~\cite{yu2023physformer++}, experiments on the MMSE-HR dataset were conducted by using 5-fold cross-validation method for heart rate estimation.
The folds were split independently between subjects to demonstrate generalizability of the proposed method.
Table~\ref{tab:MMSE-HR hr} presents the averaged performance over 5 folds for estimating heart rate from facial rPPG signals.
The proposed method achieved MAE of 1.78, RMSE of 4.27, and \textit{r} of 0.95, respectively, outperforming previous deep learning models with a significant margin.
Previous methods proposed by Nowara et al.~\cite{nowara2021benefit}, Jaiswal \& Meenpal~\cite{jaiswal2022heart}, Ouzar et al.~\cite{ouzar2023x} infer rPPG signals based on small window sizes with low latency.
However, insufficient temporal information leads to increase errors in estimating rPPG signals, resulting in higher heart rate estimation errors.
Video transformer-based models proposed by Yu et al.~\cite{yu2022physformer} and Yu et al.\cite{yu2023physformer++} demonstrate significant improvements in cross-dataset tests.
However, transformers require high-performance computing resources due to their large number of parameters and computational demands.
As shown in Table~\ref{tab:MMSE-HR hr}, our proposed model achieved better performance compared to the previous methods by using similar length of video sequences.
In addition, the heart rate estimation errors from acral rPPG signals are 1.91, 4.74, and 0.93, respectively.
DRP-Net estimates two rPPG signals with different phases and minimal errors for utilization in the blood pressure estimation stage.

\begin{table}
    \caption{Heart rate estimation results on the MMSE-HR database.}
    \centering
    \label{tab:MMSE-HR hr}
    \resizebox{0.9\linewidth}{!}{%
    \setlength{\tabcolsep}{2pt}
    \renewcommand{\arraystretch}{1.3}
    \begin{tabular}{lcccc}
        \hline
        Method & Window (s) & MAE (BPM) & RMSE & \textit{r} \\ \hline
        POS~\cite{de2013robust} & - & 5.77 & - & 0.82 \\
        DeepPhys~\cite{chen2018deepphys} & - & 4.72 & 8.68 & 0.82 \\
        Benefit of distraction~\cite{nowara2021benefit} & 2 & 2.27 & 4.90 & \underline{0.94} \\
        EfficientPhys-C~\cite{liu2021efficientphys} & - & 3.48 & 7.21 & 0.86 \\
        CAN with synthetic data~\cite{mcduff2022using} & 30 & 2.26 & \textbf{3.70} & - \\
        PhysFormer~\cite{yu2022physformer} & 6.4 & 2.84 & 5.36 & 0.92 \\
        Spatiotemporal feature~\cite{jaiswal2022heart} & 5 & 6.40 & 6.82 & \textbf{0.95} \\
        X-iPPGNet~\cite{ouzar2023x} & 2 & 4.10 & 5.32 & 0.85 \\
        PhysFormer++~\cite{yu2023physformer++} & 6.4 & 2.71 & 5.15 & 0.93 \\
        CIN-rPPG~\cite{li2023channel} & 12 & \underline{1.93} & 4.43 & \underline{0.94} \\
        Dual-TL~\cite{qian2024dual} & 12 & 2.25 & \underline{4.27} & 0.93 \\ 
        Ours & 6 & \textbf{1.78} & \underline{4.27} & \textbf{0.95} \\\hline
    \end{tabular}%
    }
\end{table}

\begin{figure}[t]
    \centering
    \includegraphics[width=0.9\linewidth]{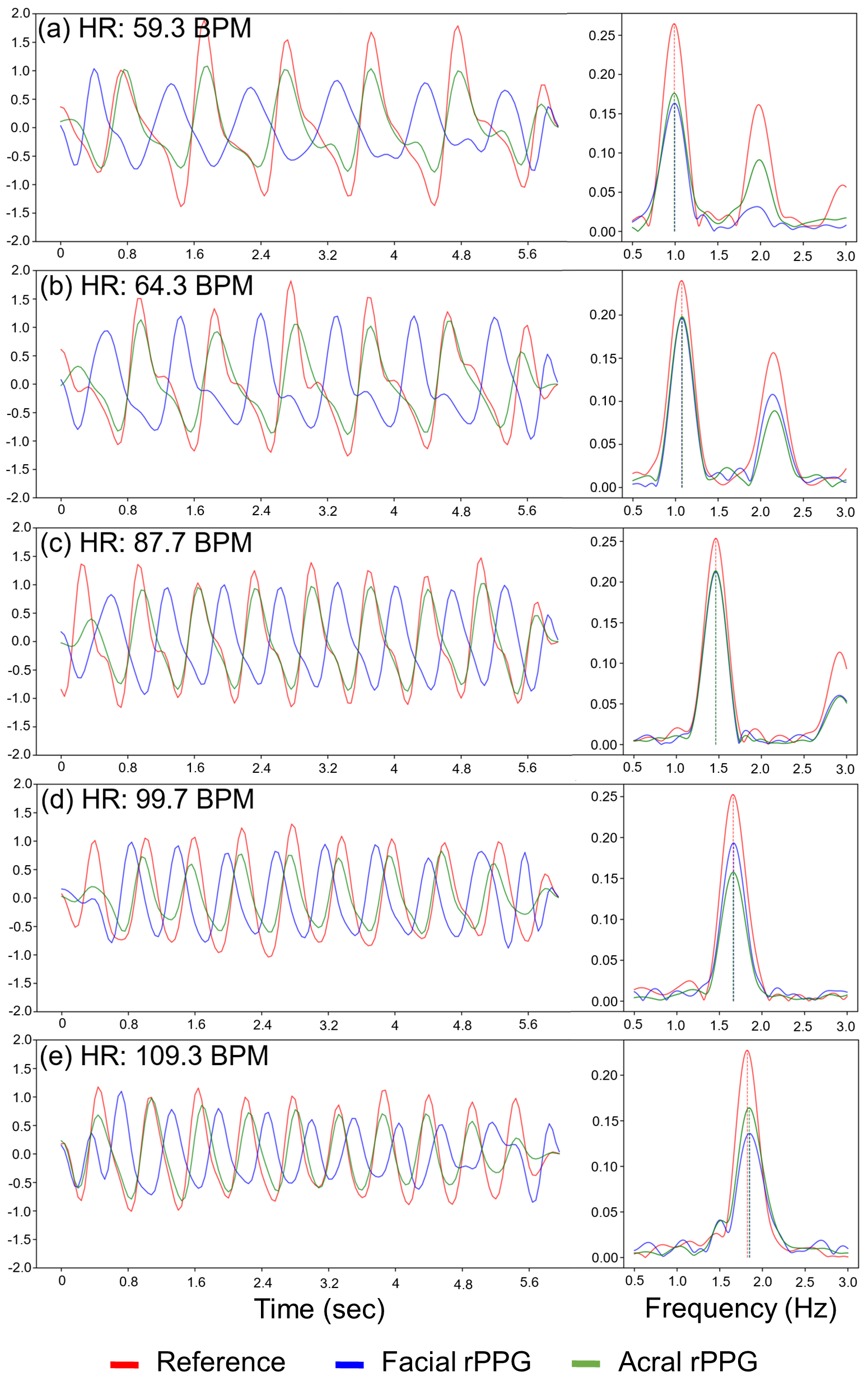}
    \caption{Visualization of rPPG signals (left) and their PSD (right).
The upper left corner of each rPPG signal displays the reference heart rate.
}
    \label{fig:Predicted_signals}
\end{figure}

Fig.~\ref{fig:Predicted_signals} shows rPPG signals and their PSD.
It is worth noting that while acral and reference rPPG signals show almost synchronized phase to each other, facial rPPG signals show phase discrepancy to the acral rPPG signals.
It implies that that different loss functions for the facial and acral rPPG signals are effective in inferring phase-shifted rPPG signals.
Figures from Fig.~\ref{fig:Predicted_signals}(a) to Fig.~\ref{fig:Predicted_signals}(e) show examples of rPPG signals in ascending order by heart rate.

Table~\ref{tab:MMSE-HR bp} presents the results of estimating blood pressure on the MMSE-HR database.
The proposed BBP-Net for blood pressure estimation outperformed previous methods, achieving MAE and RMSE of 10.19 and 13.01 for SBP and 7.09 and 8.86 for DBP, respectively.
The algorithms proposed by Rong \& Li~\cite{rong2021blood} and Schrumpf et al.~\cite{schrumpf2021assessment} extracted rPPG signals using a non-parametric approach and estimated blood pressure based on handcrafted features and deep learning models.
These previous studies are similar to our approach in the aspect of extracting rPPG signals in an intermediate step.
However, the main difference of our proposed method is to analyze phase-shifted rPPG signals extracted from DRP-Net.
Chen et al.~\cite{chen2023remote} proposed a blood pressure estimation model that utilizes two-dimensional spatiotemporal maps obtained from facial videos.
However, this previous method cannot analyze pulse transit time of rPPG signals and shows insufficient performance for estimating SBP and DBP.
Experimental results in Table~\ref{tab:MMSE-HR bp} demonstrate that analyzing temporal discrepancy in phase-shifted rPPG signals is meaningful for improving the precision of blood pressure estimation.

\begin{table}
    \caption{Blood pressure estimation results on the MMSE-HR database.}
    \centering
    \label{tab:MMSE-HR bp}
    \resizebox{1.0\linewidth}{!}{%
    \setlength{\tabcolsep}{2pt}
    \renewcommand{\arraystretch}{1.3}
    \begin{tabular}{lccclcc}
    \hline
   \multirow{2}{*}{Method} & \multirow{2}{*}{\begin{tabular}[c]{@{}c@{}}Window (s)\end{tabular}} & \multicolumn{2}{c}{SBP} &  & \multicolumn{2}{c}{DBP} \\ \cline{3-4} \cline{6-7} 
 &  & MAE (mmHg) & RMSE &  & MAE (mmHg) & RMSE \\ \hline
    NCBP~\cite{rong2021blood} & - & 17.52 & 22.43 &  & 12.13 & 15.23 \\
    NIBPP~\cite{schrumpf2021assessment} & 7 & 13.60 & - &  & 10.30 & - \\
    BPE-Net~\cite{chen2023remote} & 6 & \underline{12.35} & \underline{16.55} &  & \underline{9.54} & \underline{12.22} \\ 
    Ours & 6 & \textbf{10.19} & \textbf{13.01} &  & \textbf{7.09} & \textbf{8.86} \\ \hline
    \end{tabular}%
    }
\end{table}

\begin{figure}[t]
    \centering
    \includegraphics[width=\linewidth]{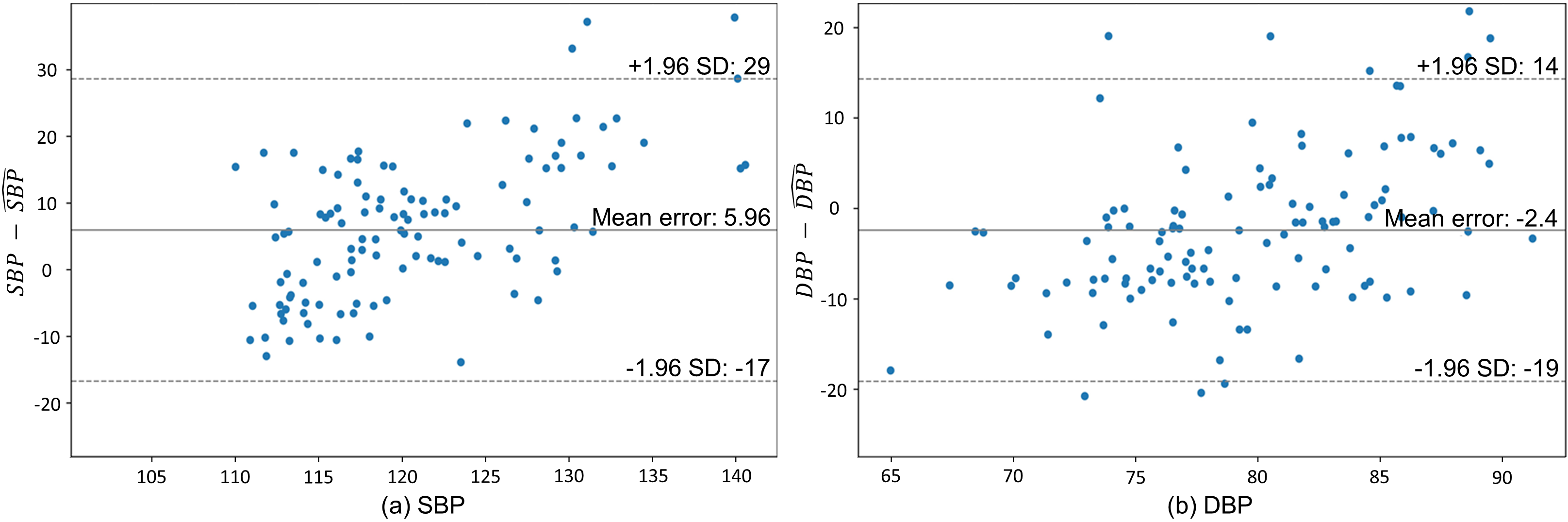}
    \caption{Bland-Altman plot of predicted and reference blood pressure on the MMSE-HR database.
The vertical axis is the signed error between predicted and reference blood pressure.
The horizontal axes in (a) and (b) indicate the reference SBP and DBP values, respectively.
}
    \label{fig:BAplots}
\end{figure}

Fig.~\ref{fig:BAplots} presents the Bland-Altman plots of predicted and reference blood pressure for the MMSE-HR database.
In Fig.~\ref{fig:BAplots}, solid line and dotted lines represent mean error and 95\% limits of agreement, respectively.
The results for SBP and DBP show positive errors at higher blood pressure and negative errors at lower blood pressure. 
These results indicate a bias in the estimated blood pressure and suggest that the generalization performance of the blood pressure estimation model is insufficient. 
To improve the generalization performance of the model, it is essential to collect more uniform dataset and broaden its distribution. 
Future work will focus on constructing datasets that include sufficient hypotensive and hypertensive data to enhance the generalizability of blood pressure estimation models.

\subsection{Experimental results on V4V database}

This section presents experimental results for estimating heart rate on the V4V database.
In Table~\ref{tab:V4V hr}, the performance of our proposed model is compared with previous methods.
Facial rPPG signals estimated from the DRP-Net outperformed others in terms of MAE and \textit{r}, with the values of 3.83 and 0.75, respectively.
The performance comparison between PhysNet~\cite{yu2019remote} and our DRP-Net indicates that utilizing wider receptive fields through atrous convolution is more advantageous for learning spatiotemporal features than the encoder-decoder structure based on three dimensional CNN.
Moreover, compared to APNET~\cite{kim2022study}, our facial rPPG signals showed lower error rates in terms of MAE and \textit{r}.
For the heart rate estimation results derived from acral rPPG signals, we achieved errors of 4.13, 10.14, and 0.73, respectively.

\begin{table}
    \caption{Heart rate estimation results on the V4V database.}
    \centering
    \label{tab:V4V hr}
    \resizebox{0.7\linewidth}{!}{%
    \setlength{\tabcolsep}{2pt}
    \renewcommand{\arraystretch}{1.3}
    \begin{tabular}{lcccc}
    \hline
    Method & Window (s) & MAE (BPM) & RMSE & \textit{r} \\ \hline
    DeepPhys~\cite{chen2018deepphys} & 30 & 10.20 & 13.25 & 0.45 \\
    PhysNet~\cite{yu2019remote} & - & 13.15 & 19.23 & \textbf{0.75} \\
    APNET~\cite{kim2022study} & - &  \underline{4.89} & \textbf{7.68} & \underline{0.74} \\ 
    Ours & 6 & \textbf{3.83} & \underline{9.59} & \textbf{0.75} \\\hline
    \end{tabular}%
    }
\end{table}

Table~\ref{tab:V4V bp} presents the results of estimating blood pressure on the V4V database.
Our proposed algorithm achieved MAE and RMSE of 13.64 and 16.78 for estimating SBP and of 9.40 and 11.90 for estimating DBP, respectively.
There is a scarcity of literature reporting on the performance of blood pressure estimation using the V4V database.
We compared the performance of the proposed method to previous algorithms proposed by Schrumpf et al.~\cite{schrumpf2021assessment} and Hamoud et al.~\cite{hamoud2023neural}.
As shown in Table~\ref{tab:V4V bp}, our BBP-Net achieved lower MAE and RMSE in estimating both SBP and DBP, with significant margins.

\begin{table}[t]
    \caption{Blood pressure estimation results on the V4V database.}
    \centering
    \label{tab:V4V bp}
    \resizebox{0.9\linewidth}{!}{%
    \setlength{\tabcolsep}{2pt}
    \renewcommand{\arraystretch}{1.3}
    \begin{tabular}{lcccccc}
    \hline
    \multirow{2}{*}{Method} & \multirow{2}{*}{Window (s)} & \multicolumn{2}{c}{SBP} &  & \multicolumn{2}{c}{DBP} \\ \cline{3-4} \cline{6-7} 
    & & MAE (mmHg) & RMSE &  & MAE (mmHg) & RMSE \\ \hline
    NIBPP~\cite{schrumpf2021assessment}& - & 31.36 & - &  & 20.62 & - \\
    Hamoud et al.~\cite{hamoud2023neural}& - & \underline{15.12} & - &  & \underline{11.17} & - \\
    Ours & 6 & \textbf{13.64} & \textbf{16.78} & \textbf{} & \textbf{9.40} & \textbf{11.90} \\ \hline
    \end{tabular}%
    }
\end{table}

\subsection{Ablation study}

\begin{table}[t]
    \caption{Comparative study using various window lengths for estimating heart rate on the MMSE-HR database.
The error rate is measured based on MAE.}
    \centering
    \label{tab:ablation-1}
    \resizebox{0.9\linewidth}{!}{%
    \setlength{\tabcolsep}{4pt}
    \renewcommand{\arraystretch}{1.3}
    \begin{tabular}{cccccccc}
    \hline
    \begin{tabular}[c]{@{}c@{}}Window (s)\end{tabular} & rPPG & Fold 0 & Fold 1 & Fold 2 & Fold 3 & Fold 4 & Average \\ \hline
    \multirow{2}{*}{2} & Facial & 11.22 & 13.71 & 9.79 & 11.96 & 15.15 & 12.37 \\
     & Acral & 12.17 & 11.31 & 10.34 & 11.73 & 14.10 & 11.93 \\ \hline
     \multirow{2}{*}{3} & Facial & 4.08 & 10.94 & 2.75 & 3.42 & 15.38 & 7.32 \\
     & Acral & 3.56 & 4.46 & 2.92 & 3.91 & 7.95 & 4.56 \\ \hline
    \multirow{2}{*}{4} & Facial & 1.86 & 2.56 & 1.90 & 2.32 & 4.26 & 2.58 \\
     & Acral & 1.79 & 2.79 & 1.95 & 2.17 & 4.05 & 2.55 \\ \hline
     \multirow{2}{*}{5} & Facial & 1.42 & 3.01 & 1.66 & 1.64 & 4.34 & 2.41 \\
     & Acral & 1.42 & 1.96 & 1.51 & 1.52 & 4.14 & 2.11 \\ \hline
    \multirow{2}{*}{6} & Facial & 1.58 & 1.08 & 0.88 & 1.79 & 3.58 & 1.78 \\
     & Acral & 1.46 & 1.00 & 0.82 & 1.97 & 4.29 & 1.91 \\ \hline
    \multirow{2}{*}{8} & Facial & 1.20 & 1.15 & 0.91 & 1.06 & 3.11 & 1.49 \\
     & Acral & 1.13 & 1.04 & 0.58 & 1.45 & 3.20 & 1.48 \\ \hline
     \multirow{2}{*}{10} & Facial & 0.95 & 0.84 & 0.50 & 0.90 & 3.25 & 1.29 \\
     & Acral & 0.88 & 0.51 & 0.66 & 0.93 & 2.88 & 1.37 \\ \hline
    \end{tabular}%
    }
\end{table}
Ablation study was conducted to analyze the effects of the heuristic parameters and components of the proposed method.
Table~\ref{tab:ablation-1} presents the results of heart rate estimation on the MMSE-HR database using different window lengths, numerically verifying the trade-off between efficiency and informativeness.
The window length of image sequences is an important heuristic parameter, which directly related with computational power and latency of deep learning models.
A longer window length increases the time required to collect input image sequences, preprocessing duration, and the model’s computational cost. 
While small window length is advantageous for reducing latency, it causes loss of temporal information, which can lead to decreased accuracy in estimating heart rate.
As shown in Table~\ref{tab:ablation-1}, MAE for estimating heart rate is improved as the window length increases in both cases of using facial and acral rPPG signals.
However, in the case of facial rPPG, the reduction in heart rate estimation error is more significant. 
These results suggest that with longer temporal information, focusing on learning the periodic patterns using $L_{freq}$ is beneficial for heart rate estimation.
When the window length was set to 2 seconds, the MAE was significantly increased because the input sequence length is smaller than the receptive field of the DRP-Net.
In experiments, we selected the window length of 6 seconds to achieve lower error rates with reduced latency compared with previous methods.

Table~\ref{tab:ablation-2} presents the ablation study on the MMSE-HR database to demonstrate the effectiveness of proposed training methods, such as data augmentation and time-domain loss $L_{pv}$.
Applying data augmentation reduced 10.05\% of MAE for estimating heart rate from facial rPPG signals from 1.99 BPM to 1.79 BPM, simultaneously reducing 15.98\% of RMSE.
This result demonstrates that augmentation of bradycardia and tachycardia data is beneficial for reducing outlier predictions of low and high heart rates.
Additionally, when applying $L_{pv}$, MAE of the DRP-Net for estimating heart rate from facial rPPG signals was reduced by 0.56\% from 1.79 BPM to 1.78 BPM, and RMSE was reduced by 5.53\% from 4.52 to 4.27.
Experimental results in Table~\ref{tab:ablation-2} demonstrate that both data augmentation and $L_{pv}$ are advantageous for improving the accuracy of estimated heart rate.

\begin{table}
    \caption{The effectiveness of data augmentation and $L_{pv}$.}
    \centering
    \label{tab:ablation-2}
    \resizebox{0.7\linewidth}{!}{%
    \setlength{\tabcolsep}{2pt}
    \renewcommand{\arraystretch}{1.3}
    \begin{tabular}{cccccc}
    \hline
    Augmentation & $L_{pv}$ & rPPG & MAE (BPM) & RMSE & \textit{r} \\ \hline
    \multirow{2}{*}{} & \multirow{2}{*}{} & Facial & 1.99 & 5.38 & 0.92 \\
    & & Acral & 2.39 & 6.97 & 0.86 \\ \hline
    \multirow{2}{*}{} & \multirow{2}{*}{\checkmark} & Facial & 2.18 & 5.92 & 0.90 \\
    & & Acral & 2.03 & 5.41 & 0.92 \\ \hline
    \multirow{2}{*}{\checkmark} & \multirow{2}{*}{} & Facial & 1.79 & 4.52 & 0.94 \\
    & & Acral & 2.04 & 5.81 & 0.91 \\ \hline
    \multirow{2}{*}{\checkmark} & \multirow{2}{*}{\checkmark} & Facial & 1.78 & 4.27 & 0.95 \\
    & & Acral & 1.91 & 4.74 & 0.93 \\ \hline
    \end{tabular}%
    }
\end{table}

\begin{figure*}[t]
    \centering
    \includegraphics[width=\linewidth]{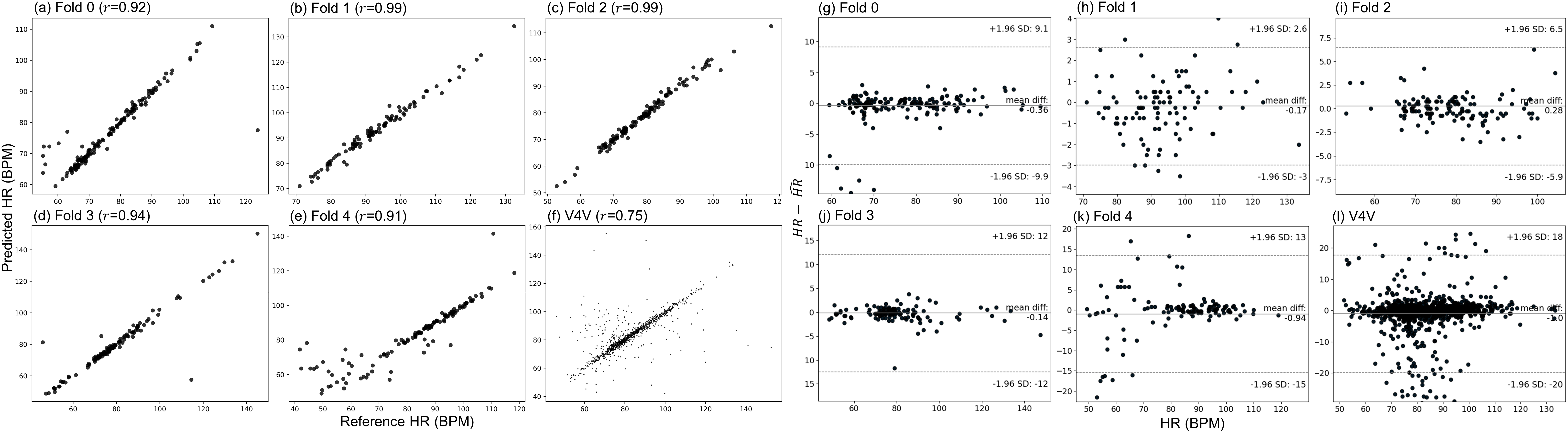}
    \caption{
    Predicted heart rate computed from facial rPPG signals and their reference heart rate.
    In subfigures (a)–(f), Pearson correlation between the predicted and reference heart rate is denoted as \textit{r}.
    Subfigures (g)–(l) show the Bland-Altman plot of predicted and reference heart rate. The vertical and horizontal axes represent the signed error between predicted and reference heart rate and the reference HR values, respectively.}
    \label{fig:Pearson_correlation}
\end{figure*}

Fig.~\ref{fig:Pearson_correlation} presents Pearson correlation and the Bland-Altman plot of the predicted and reference heart rate.
Fig.~\ref{fig:Pearson_correlation}(a) to Fig.~\ref{fig:Pearson_correlation}(e) and  Fig.~\ref{fig:Pearson_correlation}(g) to Fig.~\ref{fig:Pearson_correlation}(k) presents the results from the five folds of the MMSE-HR database.
Fig.~\ref{fig:Pearson_correlation}(f) and Fig.~\ref{fig:Pearson_correlation}(l) presents the result from the V4V database.
The predicted and reference heart rates exhibit an almost linear correlation on the MMSE-HR database, with the Pearson correlation coefficient (\textit{r}) exceeding 0.90 across all folds.
The linear correlation between the predicted and reference heart rates shows that our proposed method is accurate over the entire range of heart rate.
The proposed data augmentation method improved the estimation of both tachycardia and bradycardia data. 
However, bradycardia in fold 4 shows higher estimation errors due to the significant gap between the distributions of the original training and test datasets.
While Pearson correlation is 0.75 on the V4V database, our proposed method shows lower MAE and RMSE compared to previous methods as presented in Table~\ref{tab:V4V hr}.
The Bland-Altman plots illustrate a uniform distribution of positive and negative errors. 
The unbiased trend suggests strong generalizability of our proposed model.

\begin{table}
    \caption{The effectiveness of phase-shifted rPPG signals and scaled sigmoid function for estimating blood pressure.}
    \centering
    \label{tab:ablation-4}
    \resizebox{1\linewidth}{!}{%
    \setlength{\tabcolsep}{2pt}
    \renewcommand{\arraystretch}{1.5}
    \begin{tabular}{ccccccccc}
    \hline 
    \multirow{2}{*}{\shortstack{Facial\\rPPG}} & & \multirow{2}{*}{\shortstack{Acral\\rPPG}} & \multirow{2}{*}{\shortstack{Scaled sigmoid\\function}} & \multicolumn{2}{c}{SBP} &  & \multicolumn{2}{c}{DBP} \\ \cline{5-6} \cline{8-9} 
      &  &  &  & MAE (mmHg) & RMSE &  & MAE (mmHg) & RMSE \\ \hline
    \checkmark & & & \checkmark & 12.51 & 16.79 & & 7.93 & 9.41 \\
    & & \checkmark & \checkmark & 13.18 & 17.89 & & 7.84 & 9.42 \\ 
    \checkmark & & \checkmark & & 19.57 & 25.34 &  & 10.15 & 13.34 \\
    \checkmark & & \checkmark & \checkmark & 10.19 & 13.01 &  & 7.09 & 8.86 \\ \hline
    \end{tabular}%
}
\end{table}

To analyze the performance of BBP-Net for estimating blood pressure, ablation study was conducted to demonstrate the effectiveness of utilizing phase-shifted rPPG signals and scaled sigmoid function.
Table~\ref{tab:ablation-4} presents MAE and RMSE for estimating SBP and DBP from different types of rPPG signals.
While the MAEs for estimating SBP from facial and acral rPPG signals were 12.51 mmHg and 13.18 mmHg, it was reduced to 10.19 mmHg when utilizing both rPPG signals.
This result indicates that the temporal discrepancy in pulse waves at different physiological sites contributes to reducing the error for estimating blood pressure.
Table~\ref{tab:ablation-4} shows the performance of blood pressure estimation with and without the scaled sigmoid function in BBP-Net.
Without the scaled sigmoid function, the MAEs for SBP and DBP were increased by 9.38 mmHg and 3.06 mmHg, respectively.
These experimental results demonstrate that constraining predicted blood pressure into a bounded range is effective to reduce the error by eliminating outlier estimates.

\subsection{Cross skin tone testing}

\begin{figure}
    \centering
    \includegraphics[width=\linewidth]{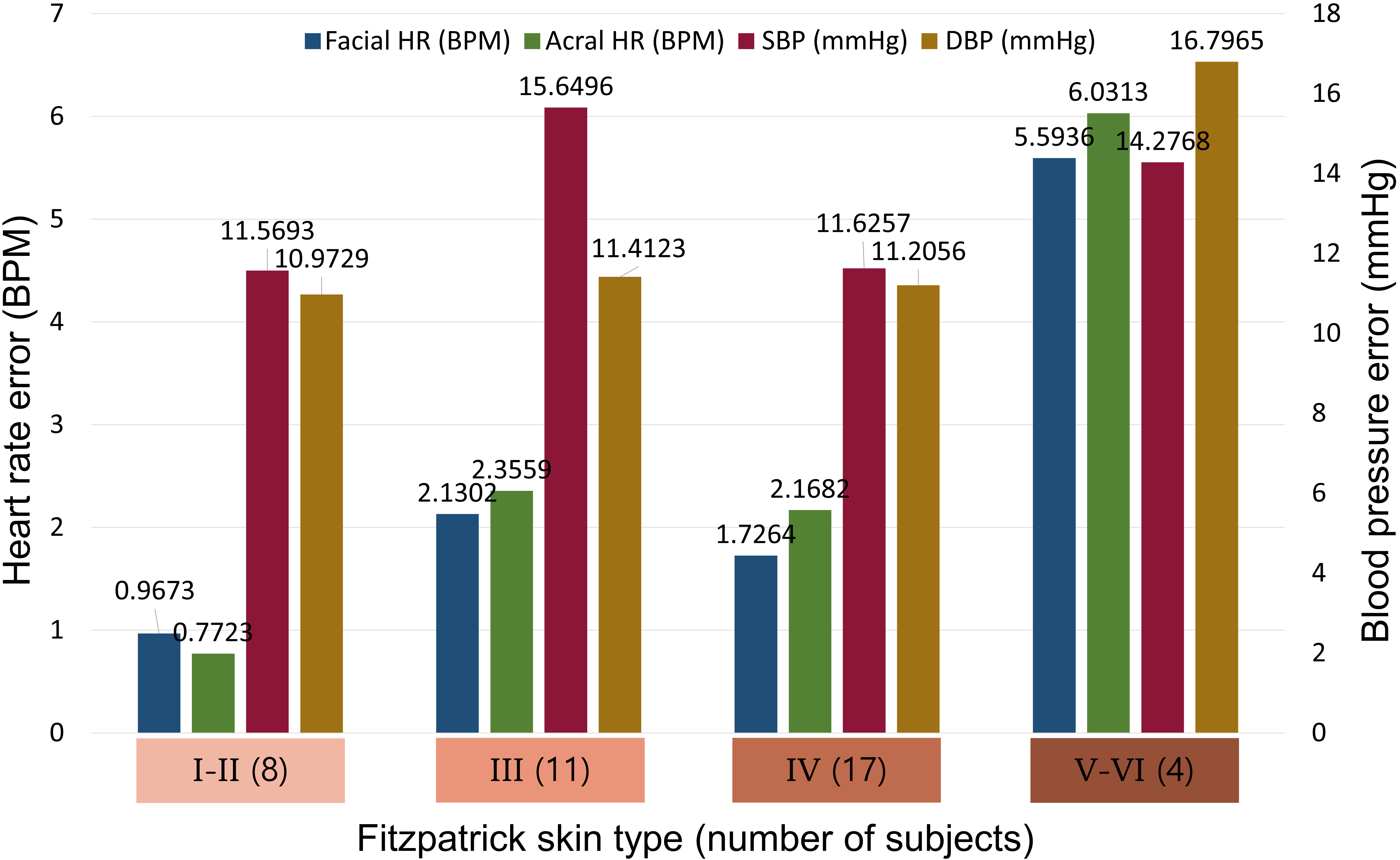}
    \caption{Cross skin tone testing. The horizontal axis represents the Fitzpatrick skin type, while the blue, green, red, and yellow bars represent the MAEs of HR from facial rPPG, HR from acral rPPG, SBP, and DBP, respectively.}
    \label{fig:cross_skin_tone_testing}
\end{figure}

Cross skin tone testing was conducted to evaluate the robustness of the proposed DRP-Net and BBP-Net across various skin tones. Skin tones were divided into four folds according to the Fitzpatrick skin type~\cite{fitzpatrick1988validity}, which were type I-II, type III, type IV, and type V-VI.
Fig.~\ref{fig:cross_skin_tone_testing} presents the cross-validation results by skin tone in the form of bar graphs, where the left and right axes represent heart rate and blood pressure errors in MAE, respectively.
The acral rPPG-based heart rate estimation error was observed to be 0.77 BPM for lighter tones and 6.03 BPM for darker tones, indicating that it is more challenging to detect periodic signals as skin tone becomes darker.
Similarly, the average SBP and DBP estimation errors were highest for skin types V-VI.
These findings suggest that the robustness of the proposed pipeline appears to decrease for darker skin tones, indicating a limitation that requires further investigation in future research.

\subsection{Computational complexity}

\begin{table}[]
    \caption{Comparison of computational cost.}
    \centering
    \label{tab:ablation-5}
    \resizebox{0.6\linewidth}{!}{%
    \setlength{\tabcolsep}{2pt}
    \renewcommand{\arraystretch}{1.5}
    \begin{tabular}{lcc}
    \hline
    Methods                      & Parameters    & MACs         \\ \hline
    PhysNet~\cite{yu2019remote}  & 0.73 M                & 65.19 G               \\
    TS-CAN~\cite{liu2020multi}   & 3.91 M                & 61.96 G               \\
    AutoHR~\cite{yu2020autohr}   & 0.99 M                & 189.22 G               \\
    EfficientPhys-C~\cite{liu2021efficientphys} & 3.84 M                & 31.32 G                \\
    PhysFormer~\cite{yu2022physformer}      & 7.03 M                & 47.01 G                \\
    PhysFormer++~\cite{yu2023physformer++}  & 9.79 M                & 49.85 G                \\
    Ours                                    & 0.74 M                & 38.11 G                \\ \hline
    \end{tabular}%
    }
\end{table}

Table~\ref{tab:ablation-5} presents the number of parameters and the multiply-accumulates (MACs) for both the previous and proposed heart rate estimation models. 
The proposed DRP-Net achieves the lowest complexity and heart rate estimation error compared to state-of-the-art methods. 
Notably, transformer-based models [14, 31] have more than 10 times the number of parameters compared to the proposed model. 
While PhysNet [30] has a similar number of parameters, it requires approximately 1.71 times more MACs. 
Additionally, the complexity of a blood pressure estimation model, which shows 0.85 M parameters and 15.6 M MACs.
Moreover, the parameters and MACs of the blood pressure estimation model are 0.85 M and 15.6 M, respectively.
The efficient 2-stage deep learning framework allows for the acquisition of multiple physiological information with minimal computational cost.

\subsection{International standard}

We analyze the performance of the proposed method for estimating blood pressure based on the international standard of the British Hypertension Society (BHS) \cite{o1990british}.
The BHS standard evaluates the percentages of estimates which satisfy the MAEs lower than 5 mmHg, 10 mmHg, and 15 mmHg.
For example, to obtain Grade A of the BHS standard, more than 60\%, 85\%, and 95\% of estimates should satisfy the MAEs lower than 5 mmHg, 10 mmHg, and 15 mmHg, respectively.
Table~\ref{tab:BHS} presents the percentages within three thresholds for MAE for estimating SBP and DBP on the MMSE-HR and V4V databases.
Our proposed method achieved Grade B on the MMSE-HR database and Grade C on the V4V database for estimating DBP.

\begin{table}
    \caption{The criteria of the BHS standard and the accuracy of estimating blood pressure based on three thresholds for MAE.}
    \centering
    \label{tab:BHS}
    \resizebox{0.8\linewidth}{!}{%
    \setlength{\tabcolsep}{2pt}
    \renewcommand{\arraystretch}{1.3}
    \begin{tabular}{lcccc}
    \hline
     &  &  $\leq$ 5 mmHg & $\leq$ 10 mmHg & $\leq$ 15 mmHg \\ \hline
    \multirow{2}{*}{MMSE-HR} & SBP & 46.02 \% & 65.49 \% & 75.22 \% \\
     & DBP & 81.42 \% & 92.04 \% & 94.69 \% \\
    \multirow{2}{*}{V4V} & SBP & 65.24 \% & 72.94 \% & 81.39 \% \\
     & DBP & 64.17 \% & 80.11 \% & 86.63 \% \\ \hline
    \multirow{3}{*}{BHS standard} & Grade A & 60 \% & 85 \% & 95 \% \\
     & Grade B & 50 \% & 75 \% & 90 \% \\
     & Grade C & 40 \% & 65 \% & 85 \% \\ \hline
    \end{tabular}%
}
\end{table}

\section{Discussion}
\label{sec:discussion}

In this paper, we aim to discover direct clues for blood pressure estimation by extracting phase-shifted rPPG signals from facial videos.
We demonstrate the superiority of the proposed deep learning framework and detailed training techniques through comparative experiments with previous methods and ablation studies.
However, validating the phases of facial rPPG signals is challenging due to the absence of ground truth PPG signals measured from the face. 
The temporal discrepancy between facial rPPG and acral rPPG is a crucial factor in improving blood pressure estimation performance, and insufficient validation could be considered a limitation of this study.

\begin{figure}[t]
    \centering
    \includegraphics[width=\linewidth]{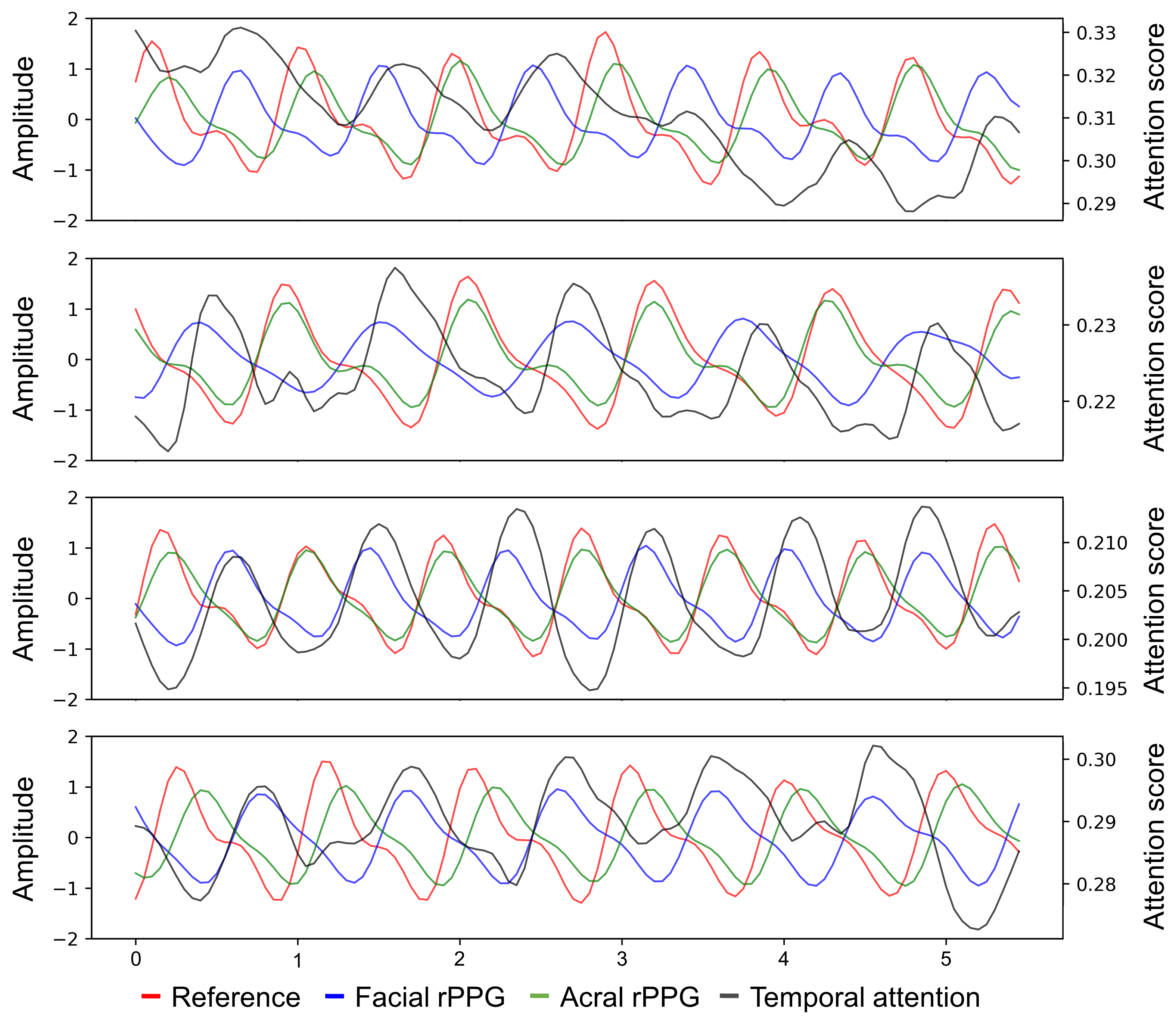}
    \caption{Visualization of temporal attention $\alpha_t$. The black, blue, green, and red signals represent temporal attention, facial rPPG, acral rPPG, and ground truth PPG signals, respectively.}
    \label{fig:temporal_attention}
\end{figure}

\begin{figure}[t]
    \centering
    \includegraphics[width=\linewidth]{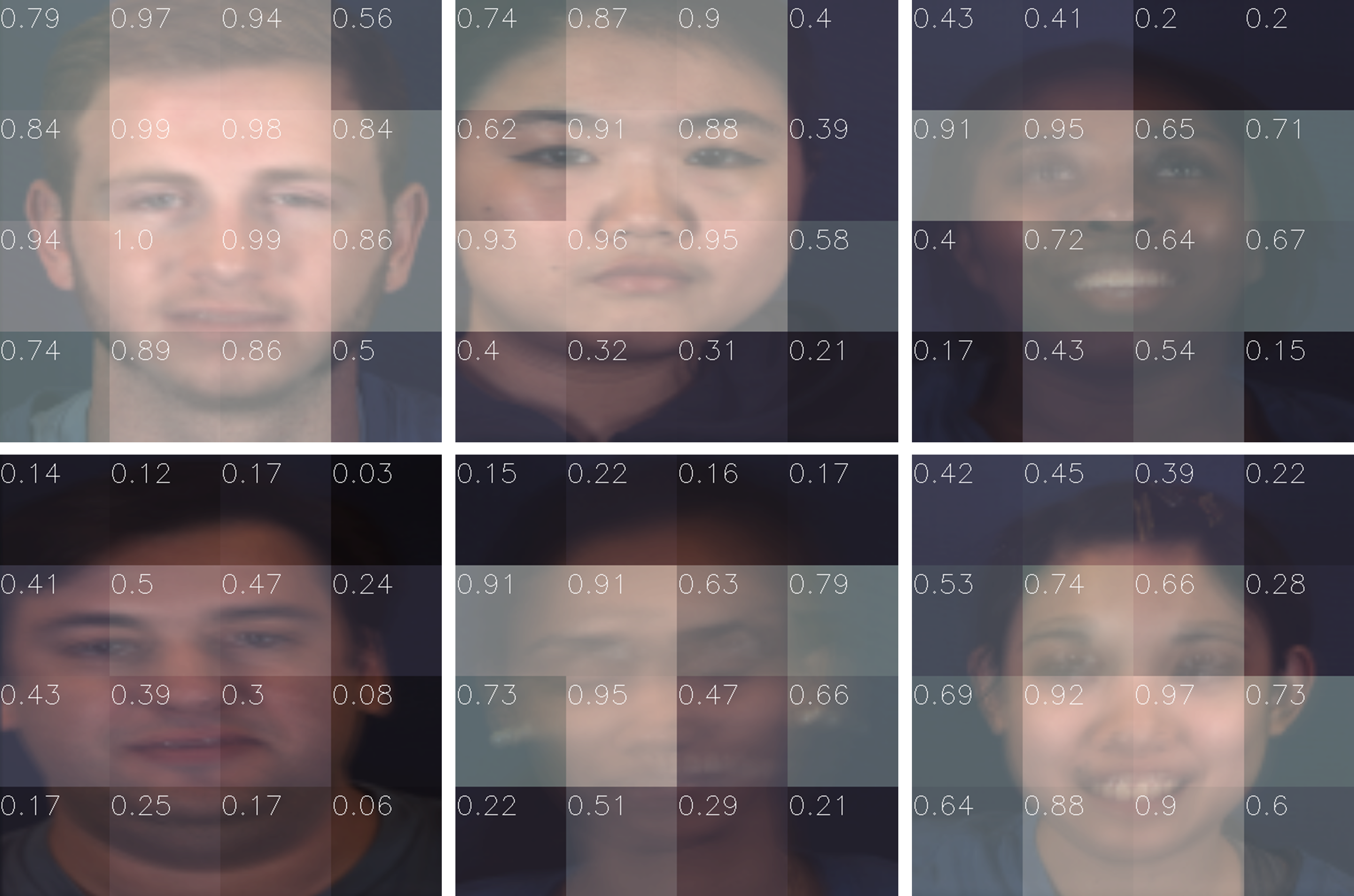}
    \caption{Visualization of spatial attention $\alpha_s$. The highlighted areas illustrate regions with higher attention scores.}
    \label{fig:spatial_attention}
\end{figure}

To address this limitation, we visualized the spatial and temporal attention of DRP-Net to identify the spatiotemporal locations where the model assigned higher weights in the frames. 
In Fig.~\ref{fig:temporal_attention}, the temporal attention exhibited a periodic pattern similar to the physiological signals, particularly with a phase closely aligned with facial rPPG.
The spatial attention was designed with a size of $4\times4$, as the spatial dimension of the feature map is reduced. Fig.~\ref{fig:spatial_attention} illustrates the visualization of spatial attention, overlaid on the clip-averaged images. 
The visualization results show that higher scores were assigned to facial regions across various skin tones. Notably, for the motion data in the second row, the score is also higher in regions where skin is mainly present.

These attention visualization results suggest that the model has learned the periodicity of emphasized subtle skin tone changes in each facial frame.
However, to reliably validate facial rPPG signals, it is necessary to measure facial PPG signals utilizing sensors attached to the face. 
Future work will focus on constructing a real dataset that includes PPG signals from various body sites, with two main objectives. 
First, verifying rPPG signals extracted from captured body sites using deep learning models. 
Second, developing a model for blood pressure estimation based on the temporal discrepancy of physiological signals across various body sites.

\section{Conclusion}
\label{sec:conclusion}
This paper proposes a two-stage deep learning framework consisting of DRP-Net and BBP-Net for estimating heart rate and blood pressure from facial videos.
In this paper, we introduce the concept of phase-shifted rPPG signals for analyzing phase discrepancy of pulse waves at facial and acral regions.
The DRP-Net extracts facial and acral rPPG signals based on 3D convolution and Siamese-structured heads, and a time domain loss is proposed to supervise the scale of the estimated rPPG signals.
The loss function in the frequency domain enabled the learning of rPPG signal phases corresponding to the video sequences of the captured body sites.
The BBP-Net estimates SBP and DBP values within a bounded range from the phase-shifted rPPG signals.
Phase-shifted rPPG signals provide insights for blood pressure estimation by conveying the time delay of the cardiac cycle.
Moreover, the frame interpolation based data augmentation method makes the heart rate distribution wider resulting reduce the bias of the model.
Experiments were conducted on the MMSE-HR and V4V databases, and our proposed method achieved superior performance with a significant margin compared to previous methods.
Ablation study was thoroughly performed to demonstrate the effectiveness of the phase-shifted rPPG signals for estimating heart rate and blood pressure.

\bibliographystyle{IEEEtranDOI}
\bibliography{reference}

\vfill

\end{document}